# Trust as Extended Control: Active Inference and User Feedback During Human-Robot Collaboration

**Felix Schoeller**, Massachusetts Institute of Technology, Cambridge, MA, United States. Gonda Multidisciplinary Brain Centre, Bar Ilan University, Israel

**Mark Miller**, Center for Human Nature, Artificial Intelligence and Neuroscience, Hokkaido, Japan; Department of Informatics, University of Sussex, UK.

**Roy Salomon**, Gonda Multidisciplinary Brain Centre, Bar Ilan University, Israel

**Karl J. Friston**, Wellcome Trust Centre for Neuroimaging, University College London, London, UK.

Manuscript type: Review

Exact word count of text: 6367

Contact information for corresponding author: felixsch@mit.edu

**Abstract**

**Objective**: To interact seamlessly with robots, users must infer the causes of a robot's behavior—and be confident about that inference (and its predictions). Hence, trust is a necessary condition for human-robot collaboration (HRC). However, and despite its crucial role, it is still largely unknown how trust emerges, develops, and supports human' relationship to technological systems.
**Background**: We review the literature on trust, human-robot interaction, human-robot collaboration, and human interaction at large. Early models of trust suggest that it is a trade-off between benevolence and competence, while studies of human-to-human interaction emphasize the role of shared behavior and mutual knowledge in the gradual building of trust.
**Method**: Here, we introduce a model of trust as an agent' best explanation for reliable sensory exchange with an extended motor plant or partner. This model is based on the cognitive neuroscience of active inference and suggests that, in the context of HRC, trust can be casted in terms of virtual control over an artificial agent. Interactive feedback is a necessary condition to the extension of the trustor's perception-action cycle.
**Results**: This model has important implications for understanding human-robot interaction and collaboration—as it allows the traditional determinants of human trust, such as the benevolence and competence attributed to the trustee, to be defined in terms of hierarchical active inference, while vulnerability can be described in terms of information exchange and empowerment. Furthermore, this model emphasizes the role of user feedback during HRC and suggests that boredom and surprise may be used in personalized interactions as markers for under and over-reliance on the system.
**Conclusion**: The description of trust as a sense of virtual control offers a crucial step towards
grounding human factors in cognitive neuroscience and improving the design of human-centered technology. Furthermore, we examine the role of shared behavior in the genesis of trust, especially in the context of dyadic collaboration, suggesting important consequences for the acceptability and design of human-robot collaborative systems.

**Keywords**: robotics, motor control, surprise, extended agency, empowerment,

**Abstract** (in paragraph form)

To interact seamlessly with robots, users must infer the causes of a robot's behavior—and be confident about that inference (and ensuing predictions). Hence, trust is a necessary condition for human-robot collaboration (HRC). Despite its crucial role, it is largely unknown how trust emerges, develops, and supports human interactions with nonhuman artefacts. Here, we review the literature on trust, human-robot interaction, human-robot collaboration, and human interaction at large. Early models of trust suggest that trust entails a trade-off between benevolence and competence, while studies of human-to-human interaction emphasize the role of shared behavior and mutual knowledge in the gradual building of trust. We then introduce a model of trust as an agent's best explanation for reliable sensory exchange with an extended motor plant or partner. This model is based on the cognitive neuroscience of active inference and suggests that, in the context of HRC, trust can be cast in terms of virtual control over an artificial agent. In this setting, interactive feedback becomes a necessary component of the trustor's perception-action cycle. The resulting model has important implications for understanding human-robot interaction and collaboration—as it allows the traditional determinants of human trust, such as the benevolence and competence attributed to the trustee, to be defined in terms of active inference, while vulnerability can be described in terms of information exchange and empowerment. Furthermore, this model emphasizes the role of user feedback during HRC and suggests that boredom and surprise may be used in personalized interactions as markers for under and over-reliance on the system. The description of trust—as a sense of virtual control—offers a crucial step towards grounding human factors in cognitive neuroscience and improving the design of human-centered technology. Furthermore, we examine the role of shared behavior in the genesis of trust, especially in the context of dyadic collaboration, suggesting important consequences for the acceptability and design of human-robot collaborative systems.

# Trust as Extended Control: Active Inference and User Feedback During Human-Robot Collaboration

## 1. Introduction

Technology greatly extends the scope of human control and allows our species to thrive by engineering (predictable) artificial systems to replace (uncertain) natural events (Pio-Lopez et al., 2016). Navigating and operating within the domain of regularities requires considerably less motor and cognitive effort (e.g., pressing a switch to lift heavy weights) and less perceptual and attentional resources (Brey, 2000); thereby increasing the time and energy available for other activities. However, the inherent complexity of nonhuman artefacts—such as robots—invariably leads to a state of "epistemic vulnerability", whereby the internal dynamics of the system are hidden to the user and, crucially, must be inferred from the observer via the behavior of the system. Indeed, current misgivings about machine learning rest upon the issue of explainability and interpretability (Doshi-Velez & Kim 2017); namely, the extent to which a user can understand what is going on "under the hood" (Došilović, et al 2018). Crucially, the opacity of these processes may give rise to suspicions and qualms regarding the agent's goals. What factors influence trust during human-robot interaction, and how does human inference underwrite the continuous information exchange in human-computer systems?

It is widely recognized that trust is a precondition to (successful) human-machine interactions (Sheridan, 2019; Lee & See, 2004). However, despite great effort from researchers in the field, we still lack a formal (i.e., computational) understanding of the role of trust in successful human interactions with complex artificial systems. Here, we review contemporary theories of trust and their associated empirical data in the context of industrial automation. Drawing on the literature in cognitive science on active inference (Friston et al., 2006), control (Sheridan, 2019), and hierarchical perception-action cycles (Salge & Polani, 2017), we introduce a cross-disciplinary framework of trust—modelled as a sense of virtual control in the context of automation and human robot interaction. To understand the role of trust in robotics, we first present a brief overview of basic cognitive functions, focusing on the organization of motor control. We then explain the fundamental components of trust—in terms of active inference—and conclude with some remarks about the emergence and development of trust in the context of dyadic human-robot collaboration.

**2. Surprise Minimizing Agents**

From the standpoint of contemporary cognitive neuroscience, perception and action are means for living organisms to reduce their surprise (i.e., acquire information) about (past, current, and future) states of the world (Friston et al., 2006). The brain, according to this framework is a constructive, statistical organ that continuously generates hypotheses (i.e., beliefs) to predict the most likely causes of the sensory data it encounters (i.e., sensations). These predictions then guide behavior accordingly in a top-down fashion (Gregory, 1980). Various unifying and complementary theories have been proposed to describe this process (e.g., the free energy principle, active inference, predictive processing, dynamic logic, and the Bayesian brain hypothesis). These proposals rest on three fundamental brain functions that can be defined as follows: 1) perception senses change in the surroundings, 2) cognition predicts the consequences of change, and 3) action controls the causes of change. This tripartition is reflected in the hierarchical functional architecture of brain systems (Kandel, 2000), speaking to the brain as an engine of prediction ultimately aiming at the minimization (and active avoidance) of surprising states (see figure 1). There are several ways of describing the requisite (neuronal) message passing—in terms of Bayesian belief updating (Friston et al 2017). Perhaps the most popular at present is predictive coding (Rao & Ballard 1999), where inference and learning is driven by prediction errors, and agency emerges from perception-action loops (Fuster 2004, Parr & Friston 2019), continuously exchanging information with the sensorium.

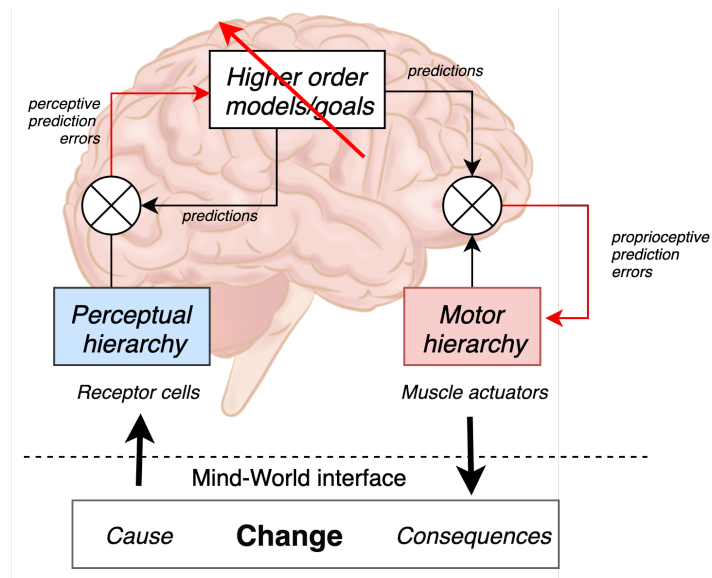

*Figure 1: Perception models afferent changes in states of the world detected by receptor cells (e.g., in the retina) and throughout the perceptual hierarchy. In this control diagram, ⊗ denotes a comparator. The red arrows denote inference and learning (i.e., driven by prediction errors) that compare (descending) predictions with (ascending) sensations. Cognition and higher order processing attempt to predict sensory input and futures states of the world based on available (generative) models; thereby, minimizing prediction error. Action organizes the motor hierarchy to actively control the efferent consequences of ongoing events; namely, by modifying causes anticipated through perceptual means, thereby altering the system' dynamics to make them more predictable (i.e., less surprising). Though not specified on this diagram, perception can be further subdivided into interoception and exteroception; respectively, modelling changes in the internal and external world. Emotion—and related notions of selfhood—usually arise via predictive processing of interoceptive sensations, often known as interoceptive inference (Seth, 2013, 2014; Seth & Friston, 2016).*

As underwriting perception and action (Méndez et al., 2014), cognition (i.e., active inference or planning) is closely related to evaluating the consequences of action in relation to prior beliefs about homeostatic needs of survival and reproduction; preparing responses to anticipated change (Pessoa, 2010). Here, beliefs correspond to Bayesian beliefs (i.e., posterior probability distributions over some hidden state of the world)—as opposed to propositional beliefs in the folk psychology sense. Minds and their basic functions—such as perception, emotion, cognition, and action—ultimately seek good predictive control. That is, they are continuously aiming to minimize uncertainty about states of the world, where uncertainty is simply expected surprise (i.e., entropy), given a course of action. There are two fundamental ways to avoid (expected) surprise: 1) change one's cognition, beliefs, or hypotheses (i.e., perception), or 2) change the world (i.e., action). This distinction is crucial in the context of robotic systems, which are quintessentially concerned with changing the causes of sensations, rather than changing perceptual inference via cognition (Jovanovic, 2019).

In short, action aims at reducing uncertainty, where exploratory behavior leads us to interact "freely" with objects in the world—to improve our generative models of the way 'objects' behave, and ultimately rendering these behaviors more predictable (Pisula & Siegel, 2005). A generative model is at the heart of active inference—and indeed the current treatment. Technically, generative models are a probabilistic specification of how (sensory) consequences are caused by hidden or latent states of the world. The generative model generally comprises a likelihood; namely, the probability of a sensory outcome given a hidden state—and prior beliefs over hidden states. Maximizing the fit or alignment between a generative model of the sensed world—and the process generating sensory outcomes corresponds to minimizing surprise (e.g., prediction error) or—in more statistical terms—maximizing the evidence for their model (Hohwy, 2016). In the setting of active inference, this is often referred to as self-evidencing. In active inference, (expected) surprise is approximated with (expected) variational free energy; thereby providing a tractable objective function for perception and action. The integration of efferent (motor) and afferent (sensory) signals results in what can be termed the sensation of control, or feeling of agency, whereby sensorimotor mismatch is minimized (Salomon, 2016).

These three functions of perception-cognition-action form a hierarchical system with sensorimotor signals at the lowest levels of the hierarchy, and abstract cognition (executive functions of goal- directed planning and decision-making) at the highest levels (Schoeller et al., 2019). Perception is organized in a hierarchical fashion, with bottom-up sensory signals (e.g., "a change in color from red to green") being continuously predicted by top-down cognitive models (e.g., "green-light authorization for crossing the street"). Action models are also organized hierarchically, whereby fine motor interaction with the external world (e.g., typing on a keyboard), are contextualized by higher order goals (e.g., writing a paragraph), themselves prescribed by high abstract plans (e.g., getting a paper accepted in a conference)—ultimately underwriting existential goals—corresponding to the organization of life itself (Schoeller et al., 2019).

A key notion—for the requisite Bayesian belief updating above—is precision. In predictive coding formulations of active inference, precision corresponds to the estimated predictability or reliability of prediction errors (the higher the precision, the more impactful the prediction errors on Bayesian belief updating). In Active Inference terms, precision represents the agent's confidence that certain action policies (i.e., sequence of actions) will produce the states the agent expects (Friston et al 2014). Predictive agents decide what actions to pursue based on the predicted sensory consequences of the action—choosing those behaviors that are most likely to minimize surprise over the long term, and so seek out familiar and unsurprising sensory states. Heuristically, precision corresponds to the rate of evidence accumulation (c.f., the Kalman gain in Kalman filtering). This means that if the rate of error reduction is faster that expected, the action policy should be deemed more precise; and if the rate is slower than expended, and errors are amassing unexpectedly, then the policy is not as successful at bringing about those future sensory states that are expected, and this should be taken as evidence that the precision is too high. In short, estimating the precision of various Bayesian beliefs and prediction errors is crucial for optimal (active) inference in an uncertain world.

Change in the rate at which error is being resolved manifests for humans as emotional valence—we feel good when error falls at a better-than-expected rate, and we feel bad when error is unexpectedly on the rise (Van de Cruys 2017; Joffily & Coricelli, 2013; Schoeller, 2015, 2016). Valence systems can therefore equip the agent with a domain general controller capable of tracking changes in error managements and adjusting precision expectations relative to those changes (Author's articles; Hesp et al. 2019). This ongoing estimation of uncertainty or precision is a reflection of an agent's perceived fitness—that is, how adaptive the agent's current predictive model is relative to their environment. We suspect that this deep relationship between precision, uncertainty and affective valence may be crucial for a formal understanding of feelings like 'trust'. Affective valence is widely acknowledged to play an important role in trust (Dunn & Schwetizer, 2005). Positive feelings have been shown to increase trust, while negative feelings diminish it (Dunn & Schweitzer 2005). The active inference framework helps to account for this evidence, suggesting that positive and negative feelings are in part a reflection of how well or poorly one can predict the consequences of action, including the actions of others. As detailed in the following section, affectivity plays a crucial role in mediating exchanges with robots, often acting as a cardinal determinant of trust in that context specifically (Broadbent et al., 2007). Therefore, robotic design that considers affect—and related higher-level constructs—are likely to enhance productivity and acceptance (Norman et al., 2003).

## 3. Agency and Empowerment in Human-Robot Interactions

The relevance of active inference for robotics has been experimentally demonstrated in (Pio-Lopez et al., 2016). In the context of automation, understanding human agency is even more important—as experimental studies have demonstrated that one can prime for agency with external cues (leading to abusive control), and clinical studies reveal that an impairment of control is associated with depression, stress, and anxiety-related disorders (Abramson, 1989; Chorpita & Barlow, 1998). The integration of efferent (motor) and afferent (sensory) signals results in what can be termed the sensation of control or a feeling of agency (Salomon et al., 2016; Vuorre, & Metcalfe, 2016), which depends on the correspondence of top-down (virtual) predictions of the outcomes of action, and the bottom-up (actual) sensations.

As illustrated in figure 1, the brain compares actual sensory consequences of the motor action with an internal model of its predicted sensory consequences. When predicted sensory consequences match incoming sensory signals, the movement is attributed to the self and a (confident) sense of agency is said to emerge (Salomon et al., 2016, Hohwy, 2007; Synofzik, 2008; Wolpert et al., 1995). Situations in which there is a mismatch between intended and observed outcomes are generally accompanied by feelings of loss of agency, and an attribution of the movement (or lack thereof) to an external source. For example, if someone was to move my arm, then there would be the sensory experience but without the prediction. If instead I was to try to move my arm, but due to anesthetic was unable to, there would be the prediction but not the sensory confirmation. On this view, agency then is just another hypothesis (or Bayesian belief) that is used to explain interoceptive, exteroceptive and proprioceptive input. If sensory evidence is consistent with my motor plans, then I can be confident that "I caused that". Conversely, if I sense something that I did not predict, then the alternative hypothesis that "you caused that" becomes the best explanation (Seth 2015). The accompanying uncertainty may be associated with negative affect such as stress or anxiety (Peters et al 2017, Stephan et al 2016). Again, the very notions of stress and anxiety are treated as higher-level constructs—that best explain the interoceptive signals that attend situations of uncertainty and adjust precision accordingly, e.g., physiological autonomic responses of the flight or fright sort (Barrett & Simmons 2015, Seth & Friston 2016).

To measure the amount of control (or influence) an agent has and perceives, Klyubin and colleagues (2005) proposed the concept of empowerment. Empowerment is a property of self-organized adaptive systems and is a function of the agent perception-action loop, more specifically the relation between sensors and actuators of the organism, as induced by interactions between the environment and the agent's morphology (Salge & Polani, 2017). Empowerment is low when the agent has no control over what it senses, and it is high the more control is evinced (Friston et al., 2006). An information-theoretic definition has been proposed, whereby empowerment is interpreted as the amount of information the agent can exchange with its environment through its perception-action cycle. Consider for example the difference between passively watching a movie and being engaged with the same content in an immersive virtual reality setting. Crucially, empowerment reflects what an agent *can* do, not what the agent actually does (Klyubin et al., 2005), and maximizing empowerment adapts sensors and actuators to each other. In other words, empowerment can be described in terms of sensorimotor fitness—i.e., the spatial and temporal relevance of the feedback the robot gets for its own behavior. For example, a robot that gets multisensory feedback on the success of its actions has greater empowerment than a robot deprived of, say, visual information or which receives delayed information (the greater the delay, the weaker the empowerment).

Technology considerably increases human empowerment (Brey, 2000), freeing the human animal from many niches or geographical constraints (e.g., climate or geology), and allowing increasingly complicated narratives and trajectories to develop within the scope of human control (e.g., cranes allow the manipulation of heavy systems beyond mere human capabilities). Predictive organisms are attracted to—and rewarded by—opportunities to improve their predictive grip on their environments—i.e., to increase their empowerment. Technological extension of the perception-action cycle offers a powerful way of expanding empowerment, but to function effectively it needs to be integrated with the agent's sensorimotor exchange with the world.

In other words, technology must engage with the agent's extended repertoire of behaviors. That inclusion requires the technology in question to be modelled internally by the agent—so that the technology becomes part the agent's sensorimotor contingencies. This (self) modelling of technological extension is key to the emergence of trust—in active inference terms: precise beliefs about how the technology will behave and evolve relative to our own behavior. This affords the possibility of extending agency beyond the realm of the body (e.g., "I am driving my car" becomes "I am driving to Liverpool"). As we attempt to show in the next section, this extension of human control beyond mere motor action and its cognitive monitoring requires trust—as a sense of virtual control in an extended perception-action cycle (Sheridan, 1988). The study of human agency has clear relevance for robotic motor control, but to our knowledge it has not yet been applied to the problem of trust in human-robot interaction. In the next section, we examine the possibility of modelling trust in relation to active inference and empowerment.

## 3. Trust as Virtual Control in Extended Agency

Within the context of human-robot interactions (Lee, 2008), optimal trust is crucial to avoid so-called disuse of technology (i.e., loss of productivity resulting from users not trusting the system), but also abuse of technology (i.e., loss of safety resulting from overreliance on the system). Hence, the cognitive neuroscience of trust has implications for both safety and management (Lee, 2008; Sheridan & Parasuraman, 2005). Indeed, technological abuse and overreliance on automation count among the most important sources of catastrophes (Sheridan & Parasuraman, 2005). From a theoretical point of view, tremendous variations exist in what trust represents and how it can best be quantified, and several definitions have been suggested with potential applications for automation (Cohen et al., 1999; Muir, 1994). An exhaustive review—of the large body of work devoted to trust literature—is outside the scope of this article: excellent reviews can be found in (Lee & See, 2004) and (Sheridan, 2019). Here, we present the fundaments of these models of trust, in the light of perception-action loops—and potential applications to robotics—to demonstrate the relevance of the active inference framework for human factors in HRI.

Several measures of trust exist in a variety of settings from management, to interpersonal, and automation. In reviewing the literature on trust, Lee and See identified three categories of definitions; all fundamentally related to uncertainty and control (2004).

The fundamental relation between trust and uncertainty appears most prescient in situations when the uncertainty derives from the realization of goals or intentions (e.g., in human-robot interactions, or employee-employer relationships), where the internal details of the agent are unknown, leaving the trustor vulnerable. In the context of robotics—where human action is extended by robotic systems—the match between goals of the (extended) human agent and those of the (extending) robotic agent is crucial in determining the success of the relation (whether the agent will make use of the robotic extension). In order of generality, the definitions identified by Lee and See are: 1) trust as intention to (contract) vulnerability, 2) trust as vulnerability, and 3) trust as estimation of an event likelihood. Note that these three general definitions, derive from early definitions of trust by Muir (1994) and Mayer et al., (1995), according to whom trust is a trade-off between ability (A) and benevolence (B), whereby a reliable system is high in both A and B (figure 3).

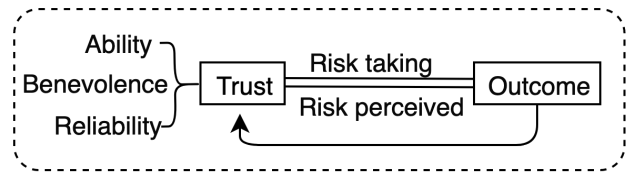

*Figure 2: Muir and Mayer model of trust as a function of the trustee's ability, benevolence, and reliability (1995) where risk perception affects risk action. This bipartition of trust as ability and benevolence amounts to two different levels in the motor hierarchy of the extended agent (e.g., the robot), whereby benevolence refers to the high-level goals motivating the extended agent and ability refers to the means of the agent to realize these goals, i.e., the sophistication of its low-level motor output in relation to the task at hand.*

The importance of externalizing goals of robotics systems (i.e., transparency) at all levels of the hierarchical perception-action loop cannot be stressed enough—for successful communication and gradual building of trust (Sheridan & Parasuraman, 2005). This is well captured in the standard definition of Sheridan (Sheridan, 2019), where communication of goals (or transparency) plays a crucial role among the seven item scales of trust.

Table 1. Standard definition of trust by Sheridan (2019)

(1) Statistical reliability (lack of error),
(2) Usefulness (ability of the system to do what is most important, e.g., in trading benefits and costs),
(3) Robustness (ability and flexibility of the system to perform variations of the task),
(4) Understandability (transparency of the system in revealing how and why it is doing what it is doing),
(5) Explication of intent (system communicating to the trustee what it will do next),
(6) Familiarity (to the user based on past experience,
(7) Dependence (upon the system by the trustee as compared to other ways of doing the given task).

In summary, trust is fundamentally related to human control to the extent that it is required for any extension of the perception-action cycle (i.e., when the success of the performance depends on some other agent's perception-action cycle, rather than one's own). Above, we saw that vulnerability is a function of empowerment in the extended agent (the more extended the agent, the more vulnerable), which can be evaluated through interaction with the robotic perception-action cycle. This may help to explain why operator curiosity is an important source of accidents in the robot industry (Lind, 2009), as curiosity aims to reduce uncertainty about the technology and so increase trust and control and suggests potential solutions in the field of accidentology. Trust is required in situations of uncertainty; and it varies as the system exhibits predictable regularities. Sheridan and Meyer models suggest that one will trust a predictable system, to the extent that one can act upon that system to obtain similar results over time, and eventually render its behavior more predictable through incremental alterations.

We have considered how a sense of agency emerges, as the resolution of mismatch between 1) the (perceptual) expectation (i.e., hypothesis) about the consequences of (motor) action, and 2) the perceived results of action (observation, perception).We introduced the idea of trust as a sense of virtual, extended control. In other words, trust is a measure of the precision, or confidence, afforded by action plans that involve another (i.e., of the match between one's actions—and their underlying intentions—and the predicted sensory consequences of another's actions). As such, 'trust' is an essential inference about states of affairs; in which the anticipated consequences of extended action are realized reliably. From the point of view of 'emotional' inference (Smith et al 2019), trust is therefore the best explanation for a reliable sensory exchange with an extended motor plant or partner. Given the role that affect plays in tuning the precision of beliefs about action policies, 'reliable' here means a predictable way to reduce expected free energy (via the extended interaction). We are attracted by, or solicited to use, a tool or device because it affords to us a means of reducing error, in a better-than-expected way, relative to doing the same work in the absence of the technological extension.

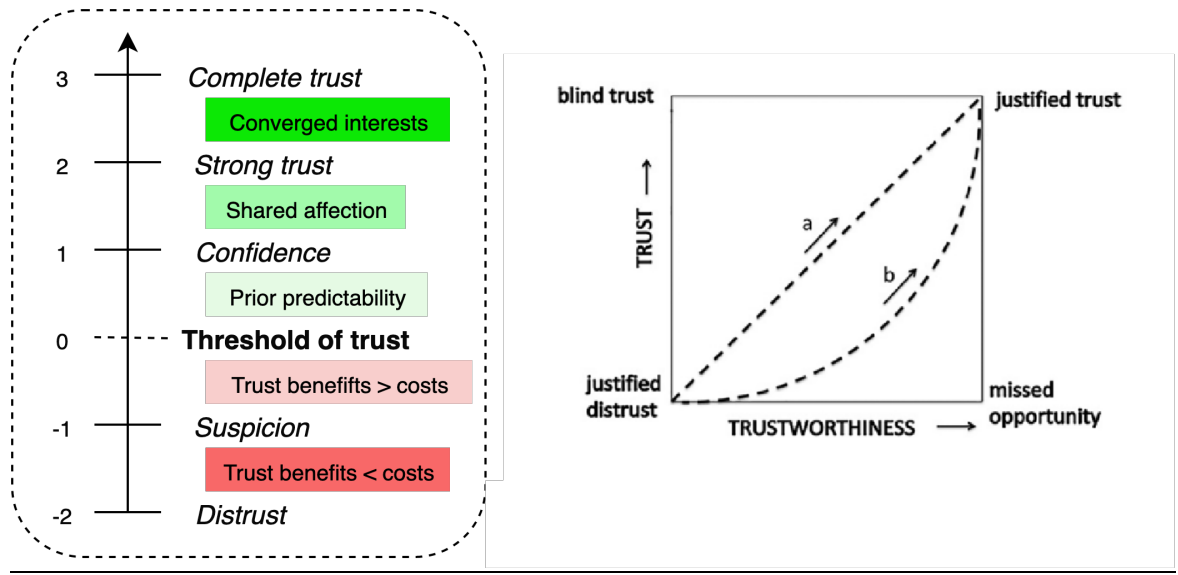

*Figure 3: On the left, levels of trust from (Dietz & Den Hartog, 2006). On the right, cross-plot of (objective) trustworthiness compared to (subjective) trust by Sheridan (Sheridan, 2019; Sheridan, 1988). As a pioneer in the study of trust in technology, Sheridan further suggested that (subjective) trust can be cross plotted against (objective) trustworthiness. This representation engenders four extremes: justified trust or distrust, blind trust (trusted untrustworthy, i.e., misuse) and missed opportunity (untrusted trustworthy, i.e., disuse). The dotted curve represents calibration, which is linear when trust is justified. Poor calibration can lead to loss of safety (due to overconfident misuse), or loss of productivity (due to underconfident disuse).*

It is generally assumed that trust in any system increases with evidence of that system's predictability or reliability (figure 3). The greater the consilience of behavioral models employed by trustor and trustee (i.e., the larger the benevolence), the greater the trust in the relationship (Hisnanick, 1989). Perhaps, this explains why simple mimicry facilitates adoption, or why one tends to agree with people who behave like ourselves—we infer shared goals based on shared behavior (Cirelli, 2018). The similarity-attraction hypothesis in social psychology predicts that people with similar personality characteristics will be attracted to each other (Morry, 2005). Hence, technology that displays personality characteristics—like those of the user—tends to be accepted more rapidly (Nass et al., 1995). As machines become increasingly intelligent, it is crucial that they communicate higher-order goals accordingly (Sheridan, 2019b). Communication of goals could therefore be optimized by rendering the perception-action cycle explicit and augmenting sensors to indicate their perceptual range (e.g., the human retina affords some information about the portion of the visual field it senses); thereby, greatly reducing the risk of accidents.

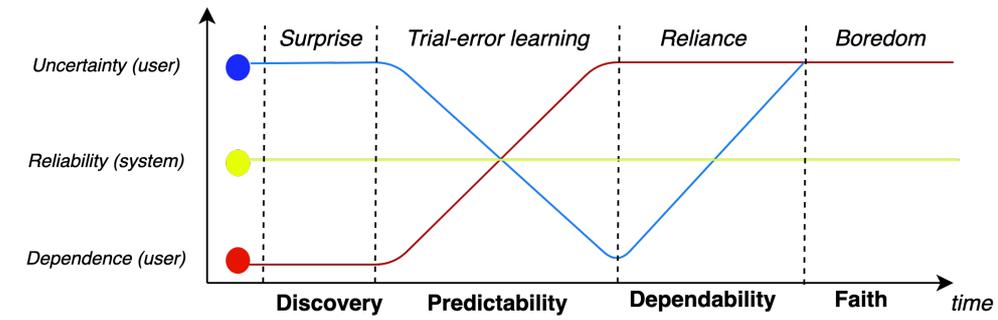

*Figure 4: Dynamics of trust over time—with four phases from discovery to faith: for a consistently reliable system, dependence (i.e., risk) is inversely proportional to uncertainty, assessed through a cycle of trial and error, until a threshold is reached. Through cycles of trial and errors, trust gradually evolves from predictability (model) to dependability (control) to a state of faith (overreliance). Our model suggests that boredom is a marker of overreliance.*

Finally, trust is a fundamentally dynamic process that eventually leads to a state of dependence. This is best exemplified in the context of information technology, whereby the information is no longer stored internally (e.g., phone numbers, navigation pathways, historical facts) but all that is known is the access pathway (my phone's contact list, my preferred web mapping service, a Wikipedia page). This is sometimes known as extended cognition, in which cognitive faculties are downloaded into the environment (Clark and Chalmers, 1998).

As suggested by the Sheridan scale, the dynamics of trust go beyond mere predictability and ultimately lead to a state of prosthetic dependence. This is evident in the context of automation, which increases the perception-action cycle at an exponential rate, thereby leading to a high abandon rate of past practices, as new technologies are adopted. Formally speaking, as technology allows the agent to reduce prediction error (by better understanding the problem space, and through more empowered actions) the agent comes to expect that rate of error reduction. The result is a gradual loss of interest or solicitation by previous less potent forms of practice.

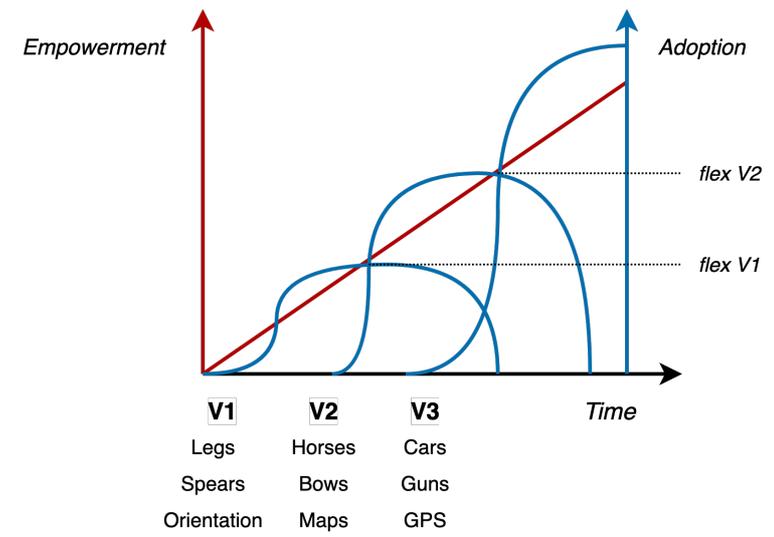

*Figure 5. Waves of technological adoption related to predictive slopes of extended engagement (empowerment) during versioning of the technology. Indeed, this is an oversimplification for the sake of visualization as we are assuming a linear progression of empowerment over time in the evolving versions of the technology (i.e., a healthy research and development cycle) where, for most technologies, newer versions may not present much greater empowerment as compared to older ones. The idea here is the inflection point (flex) indicates the start of technological decay, as the experience of better predictive rates lead to disengagement of non-extended approaches (e.g., cars replace horses replacing legs). Old practices reduce predictability at a lower rate and are rendered unsatisfactory, compared to new ones.*

In the context of interpersonal relationships, Rempel et al. (Rempel et al., 1985) describe trust as an evolving phenomenon, where growth is a function of the relationships progress. They further argue that the anticipation of future behavior forms the basis of trust at the earliest stages of a relationship. This is followed by dependability, which reflects the degree to which behavior is consistent. As the relationship matures, the basis of trust ultimately passes the threshold of faith, which has been related to benevolence (Lee & See, 2004); i.e., coordination of higher order goals that underwrite behavior. Crucially, an early study of the adaptation of operators to new technology demonstrated a similar progression (Hisnanick, 1989). Trust in that context depends on trial-and-error experience, followed by understanding of the technology's operation, and finally, a state of certainty or faith (see figure 5). Lee and Moray (Lee & Moray, 1992) made similar distinctions in defining the factors that influence trust in automation.

## 4. Trust During Dyadic Collaboration

We have seen that the essential components of trust (benevolence and competence) can be cast in terms of the confidence in beliefs at (respectively) high and low levels in the motor hierarchy, but how can active inference contribute to the science of extended agency? In this section, we examine the role of expectations in the context of dyadic interaction. So, what would a formal (first principles) approach like active inference bring to human-robot collaboration? At its most straightforward, trust is a measure of the confidence that we place in something behaving in beneficial ways that are predictable. Technically, this speaks to the encoding of uncertainty in generative models of dyadic interactions. These generative models are necessary to make inferences about policies; namely, ordered sequences of action during dyadic exchanges (Friston and Frith, 2015; Moutoussis et al., 2014).

This could range from turn taking in communication (Ghazanfar and Takahashi, 2014; Wilson and Wilson, 2005) to skilled interactions with robotic devices. At its most elemental, the encoding of uncertainty in generative models is usually framed in terms of the precision (i.e., inverse variance) or predictability (Friston et al., 2014). Crucially, every (subpersonal) belief that is updated during active inference can have the attribute of a precision or confidence. This means that the questions about trust reduce to asking what kind of belief structure has a precision that can be associated with the construct of 'trust'. In generative models based upon discrete-state spaces (e.g., partially observed Markov decision processes) there are several candidates for such beliefs. Perhaps the most pertinent—to dyadic interactions—are the beliefs about state transitions, i.e., what happens if I (or you) do that. For example, if I trust you, that means I have precise Bayesian beliefs about how you will respond to my actions. This translates into precise beliefs about state transitions during controlled exchanges (Parr et al., 2018a; Parr and Friston, 2017). This means that I can plan deep into the future before things become uncertain and, in turn, form precise posterior beliefs about the best courses of action, in other words our policies align (see figure 5).

Conversely, if I do not trust you, I will have imprecise beliefs about how you will respond and will only be able to entertain short term plans during any exchange. Furthermore, it will be difficult to infer precise outcomes of any course of action—and hence hard to entertain a shared policy. This means I will also be uncertain about which is the best course of action. Technically, this results in an imprecise belief distribution over policies, which is normally associated with negative affect or some form of angst (Badcock et al., 2017; Peters et al., 2017; Seth and Friston, 2016). Notice, that now there is not just error in the environment to deal with but also the uncertainty about another's behavior. As uncertainty increases, negatively valenced feelings emerge as a reflection of that change, and in turn reduce precision of collaborative policies. The result is the agent is less likely to enact policies of extension with that other person or robot, and so much more likely to revert to using more habitual (and relatively precise) ways of reducing error. In short, almost by definition, engaging with an untrustworthy partner is, in a folk psychological sense, rather stressful.

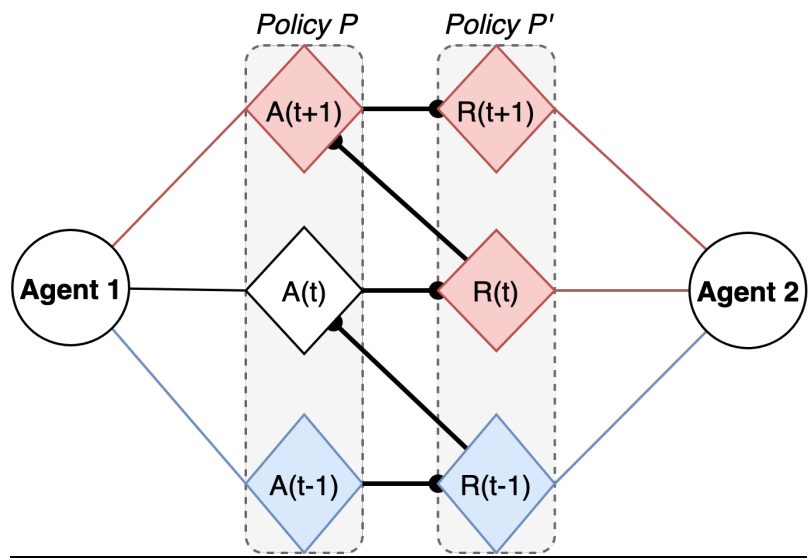

*Figure 6. A trust dyad, whereby Agent 1 performs action A at t, within the action policy P, and in collaboration with Agent 2. In a trustworthy relationship, Agent 1 can expect from Agent 2 an action Policy P', where P' is symmetrical to P (each action of P' at t+1 is a response to P at t). Past (observed) actions are blue, and future (anticipated) actions are red. The bold line in between policies represents the shared policy (or joint narrative), whereby A(t+1) can be prepared based on beliefs about anticipated R(t).*

Clearly, this active inference formulation is somewhat hypothetical. There will be many other belief structures that could be imprecise; for example, prior beliefs about the policies I should entertain and, indeed, the precision of likelihood mappings (that map from latent or hidden states of the world to observed outcomes). The latter is usually considered in terms of ambiguity (Friston et al., 2017; Veissière et al., 2019). In other words, I could consider your behavior or responses ambiguous—and that could render you untrustworthy; even if I have very precise beliefs about the latent states you are likely to navigate or pursue. In short, it may be an open question as to whether the precision of state transitions, likelihood contingencies or prior beliefs about policies manifest as differences in trust. This brings us to a fundamental motivation for the formalization of trust in terms of active inference.

It is possible to build models of dyadic exchange under ideal Bayesian assumptions using active inference, e.g., (Friston and Frith, 2015; Moutoussis et al., 2014). This means that one can optimize the prior beliefs inherent in these models to render observed choice behavior the most likely. Put another way, one can fit active inference models to empirical behavior to estimate the prior beliefs that different subjects evince through their responses (Parr et al., 2018b; Schwartenbeck and Friston, 2016). These estimates include a subject's prior beliefs about the precision of various probability distributions or Bayesian beliefs. In turn, this means it should be possible to phenotype any given person in an experimentally controlled (dyadic) situation and estimate the precision of various beliefs that best explain their behavior.

One could, in principle, then establish correlations between different kinds of precision and other validated measures of trust, such as those above. This would then establish what part of active inference best corresponds to the folk psychological—and formal definitions of trust. Interestingly, this kind of approach has already been considered in the context of computational psychiatry and computational phenotyping; especially in relation to epistemic trust (Fonagy and Allison, 2014). Epistemic trust is a characteristic of the confidence placed in someone as a source of knowledge or guidance. Clearly, this kind of trust becomes essential in terms of therapeutic relationships and, perhaps, teacher pupil relationships. Finally, one important determinant of the confidence placed in—or precision afforded—generative models of interpersonal exchange is the degree to which I can use myself as a model of you. This speaks to the fundamental importance of a shared narrative (or generative model) that underwrites any meaningful interaction of the sort we are talking about. This can be articulated in terms of a generalized synchrony that enables a primitive form of communication or hermeneutics (Friston and Frith, 2015). Crucially, two agents adopting the same model can predict each other's behavior, and minimize their mutual prediction errors. This has important experimental implications, especially in the context of human-robot collaboration, where robotic mimicry can be seen as mere self-extension for the user, leading to what philosophers of technology call relative transparency (where whatever impacts the robot also impacts me—see Brey, 2000). The self being the product of the highest prediction capacities, when another agent becomes more predictable it also increases the similarity at the highest levels in the cognitive hierarchy and thereby facilitates joint action.

This mutual predictability is also self-evident in terms of sharing the same narrative, e.g., language. In other words, my modelling of you is licensed as precise or trustworthy if, and only if, we speak the same language. This perspective can be unpacked in many directions; for example, in terms of niche construction and communication among multiple conspecifics (in an ecological context) (Constant et al., 2018; Constant et al., 2019; Veissière et al., 2019). It also speaks to the potential importance of considering self-models in human-robot collaboration design, allowing both users and robots to represent each other's behavior efficiently. Indeed, on the above reading of active inference, such shared narratives become imperative for trustworthy exchanges and collaboration. Indeed, current models suggest that the rise of subjectivity and the "self" are grounded in privileged predictive capacities regarding the states of the organism compared to the external environment (Allen & Friston, 2016; Apps & Tsakiris, 2014; Limanowski & Blankenburg, 2013; Salomon, 2017). As such, dyadic trust in another agent (biological or artificial) can be viewed as a process of extending these predictive processes beyond the body and rendering the external agent as part of a self-model. Moreover, recently robotic interfaces have been used to induce modulations of self-models by interfering with sensorimotor predictions. This in turn gives rise to phenomena closely resembling psychiatric symptoms (Blanke et al., 2014; Faivre et al., 2020; Salomon et al., 2020).

**5. Conclusion**

In the light of our increasing dependence on technology, it is worth considering that the largest aspect of human interactions with machines (i.e., their use) essentially rest upon mental models of the underlying mechanisms (e.g., few smartphone users can understand the functioning of a computer operating software). Technically, in active inference, the use of simplified generative models (e.g., heuristics) is an integral part of self-evidencing. This follows because the evidence for a generative model (e.g., of how a smartphone works) can be expressed as accuracy minus complexity. In this setting, complexity is the divergence between posterior and prior beliefs—before and after belief updating. This means the generative model is required to provide an accurate account of sensory exchanges (with a smartphone) that is as simple as possible (Maisto et al., 2015). In short, the best generative model will be, necessarily, simpler than the thing it is modelling. This principle holds true of technology in general, and automation specifically. We have examined the concept of trust from the standpoint of control and perception-action loops and found that trust components (i.e., competence and benevolence) are best casted in terms of an action-cognitive hierarchy. By examining trust from the standpoint of active inference, we were also better able to understand phenomena, such as exploration related accidents, and the gradual building of trust with shared goals, narratives, and agency. The science of human-robot interaction could make rapid progress if objective measures of trust were developed, and the neuroscience of agency does offer such metrics. It is here that a simulation setup of the sort offered by active inference could play an important part. Among the potential biomarkers for agency and control, the N1 component of event related electrical brain responses—a negative potential occurring approximately 100 ms after stimulus onset—is attenuated during self-produced or predicted events, relative to that observed during externally generated feedback. As machine become increasingly intelligent, it is to be expected that not only users will develop more sophisticated (generative) models of their internal behavior and the reliability of these behavior, but robots will also adapt to interindividual differences (Sheridan, 2019b), hence reciprocally monitor the trustworthiness of users, and thereby allow for safer and more productive interaction.

**Key points:**
- Mind— or brain—is a constructive, statistical organ that continuously generates hypotheses to predict the most likely causes of its sensory data.
- We present a model of trust as the best explanation for a reliable sensory exchange with an extended motor plant or partner.
- User boredom may be a marker of overreliance.
- Shared narratives, mutual predictability, and self-models are crucial in human-robot interaction design and imperative for trustworthy exchanges and collaboration.
- Generalized synchrony enables a primitive form of communication.
- Shared generative models may allow agents to predict each other more accurately and minimize their prediction errors or surprise, leading to more efficient human-robot collaboration.